\documentclass[conference]{IEEEtran}
\IEEEoverridecommandlockouts
\usepackage{cite}
\usepackage{amsmath,amssymb,amsfonts}
\usepackage{algorithmic}
\usepackage{graphicx}
\usepackage{textcomp}
\usepackage{xcolor}

\def\BibTeX{{\rm B\kern-.05em{\sc i\kern-.025em b}\kern-.08em
    T\kern-.1667em\lower.7ex\hbox{E}\kern-.125emX}}
\begin{document}

\title{Integrating Legal and Logical Specifications in Perception, Prediction, and Planning for Automated Driving: A Survey of Methods\\

}


\author{Kumar Manas$^{1,2 *}$, Mert Keser$^{2,3*}$, and Alois Knoll$^{3}$
    \thanks{\hspace{-5.5mm} *Authors contributed equally.}
\thanks{\hspace{-5.5mm} $^{1}$Department of Computer Science and Mathematics, Freie Universität Berlin, Germany {\tt\small kumar.manas@fu-berlin.de}}%
\thanks{\hspace{-5.5mm} $^{2}$Continental Automotive, Germany}%
\thanks{\hspace{-5.5mm} $^{3}$Computation, Information and Technology, Technical University of Munich, Germany {\tt\small mert.keser@tum.de, k@tum.de}}%
}

\maketitle

\textbf{Abstract:}
\begin{abstract}
This survey provides an analysis of current methodologies integrating legal and logical specifications into the perception, prediction, and planning modules of automated driving systems. We systematically explore techniques ranging from logic-based frameworks to computational legal reasoning approaches, emphasizing their capability to ensure regulatory compliance and interpretability in dynamic and uncertain driving environments. A central finding is that significant challenges arise at the intersection of perceptual reliability, legal compliance, and decision-making justifiability. To systematically analyze these challenges, we introduce a taxonomy categorizing existing approaches by their theoretical foundations, architectural implementations, and validation strategies. We particularly focus on methods that address perceptual uncertainty and incorporate explicit legal norms, facilitating decisions that are both technically robust and legally defensible. The review covers neural-symbolic integration methods for perception, logic-driven rule representation, and norm-aware prediction strategies, all contributing toward transparent and accountable autonomous vehicle operation. We highlight critical open questions and practical trade-offs that must be addressed, offering multidisciplinary insights from engineering, logic, and law to guide future developments in legally compliant autonomous driving systems.

\end{abstract}


\section{Introduction}
\label{sec:intro}
As autonomous vehicles (AVs) approach widespread deployment, ensuring legal compliance alongside technical performance becomes critical. Beyond generating optimal maneuvers, AV decision-making algorithms must uphold traffic regulations and provide legally defensible explanations for their actions, particularly in accident scenarios where decisions like overtaking or yielding require justification under existing law. This necessity has driven research into methodologies that integrate legal reasoning with predictive capabilities, enabling systems that balance performance and compliance.

This survey targets researchers and practitioners at the intersection of autonomous driving, formal methods, and legal compliance. We systematically reviewed literature from IEEE Xplore, ACM Digital Library, arXiv, and key venues (IV, ITSC, ICLP) using keywords including ``autonomous driving," ``formal methods," ``traffic regulation," ``rule-compliant prediction," and ``legal compliance." 

We organize existing approaches following the standard AV system architecture, covering four core areas: (1) \textbf{perception systems} as the foundation for lawful interpretation of the environment (Section~\ref{sec:perception}); (2) \textbf{formalizing traffic laws} for machine reasoning, including logic-based specifications and rule extraction from legal text (Section~\ref{sec:formalization}); (3) \textbf{legal compliance in motion planning}, encompassing rule-constrained planning and verification (Section~\ref{sec:planning}); and (4) \textbf{handling ambiguities and exceptions} in legal rules, including norm-aware behavior prediction, rule exceptions, and jurisdictional variability (Section~\ref{sec:ambiguity}). Section~\ref{sec:explain} discusses cross-cutting concerns of legal interpretability and explainability, and Section~\ref{sec:discussion} outlines open challenges and future directions.

\section{Perception Systems}
\label{sec:perception}

Reliable and interpretable perception systems are foundational to automated driving, serving as the primary interface between the vehicle and its environment. The integration of legal and logical specifications into perception presents unique challenges that extend beyond traditional computer vision paradigms. This section examines methods that enhance perception systems through neural-symbolic integration, uncertainty quantification, and formal verification—approaches that are essential for developing legally compliant and trustworthy automated driving systems.

\subsection{Neural-Symbolic Integration for Robust Perception}

A critical vulnerability in neural network-based perception systems is their susceptibility to adversarial attacks, where subtle input perturbations can cause significant classification errors with severe safety implications such as misclassifying a stop sign as a speed limit sign. Several approaches have emerged to address this fundamental challenge by integrating symbolic reasoning with neural networks.

\subsubsection{Energy-Based Modeling Frameworks}

Recent advances in neural-symbolic integration have led to unifying mathematical frameworks that combine neural perception with logical reasoning~\cite{dickens2024mathematical}. Dickens et al. propose Neural-Symbolic Energy-Based Models (NeSy-EBMs) that incorporate domain knowledge into perception systems through energy functions that score the compatibility between predictions and logical constraints~\cite{dickens2024mathematical}. This approach establishes three distinct modeling paradigms: Deep Symbolic Variables (DSVar), where neural outputs represent variables scored by symbolic logic; Deep Symbolic Parameters (DSPar), where neural modules produce symbolic parameters; and Deep Symbolic Potentials (DSPot), where neural modules generate entire symbolic potentials.
The NeSy-EBM framework is particularly valuable for autonomous driving perception as it enables the enforcement of domain constraints—such as traffic rules and physical laws—directly into the learning process, improving robustness in challenging scenarios including adverse weather conditions and partial occlusions.

\subsubsection{Logic-Infused Deep Learning}

For traffic sign recognition, the Robust Logic-infused Deep Learning (RLDL) approach enhances neural network trustworthiness by incorporating logical constraints derived through Inductive Logic Programming (ILP)~\cite{chaghazardi2024trustworthy}. This approach automatically extracts logical rules based on high-level features like shape and color, transforming them into a matrix of logical constraints that are then incorporated into the neural network's loss function.
The RLDL~\cite{chaghazardi2024trustworthy} methodology demonstrates how perception systems can leverage domain knowledge in a structured manner:
A symbolic framework extracts logical rules from human knowledge using ILP
These rules are mapped to logical constraints
A deep learning framework processes traffic sign images while ensuring compliance with the constraints through regularization
Experimental results on the German Traffic Sign Recognition Benchmark (GTSRB)~\cite{stallkamp2012man} show that RLDL models significantly outperform baseline CNNs when subjected to adversarial attacks while maintaining comparable performance on normal images. This demonstrates the potential of neural-symbolic integration to enhance robustness without sacrificing performance.

\subsubsection{T-norm-Based Constraint Integration}

Stoian et al. address the scalability challenges of incorporating logical constraints into deep learning models through memory-efficient t-norm-based loss functions~\cite{stoian2024exploiting}. Their approach enables the integration of complex logical constraints into perception systems using sparse matrix representations that dramatically reduce memory requirements from over 100 GiB to less than 25 GiB, making implementation feasible on modern GPUs.
Such logic-based approaches are particularly beneficial when training data is limited. For instance, Gödel t-norm~\cite{stoian2024exploiting} implementations have shown improvements of up to 3.95\% in mean average precision with only 20\% labeled data, addressing a key challenge in autonomous driving where obtaining fully labeled datasets is resource-intensive.

\subsubsection{Multi-Label Detection with Constrained Loss}

MOD-CL (Multi-label Object Detection with Constrained Loss) extends traditional object detection frameworks to support multiple labels per bounding box while ensuring outputs satisfy predefined logical requirements~\cite{moriyama2024mod}. This approach is particularly relevant for autonomous driving, where objects require both class labels (car, pedestrian) and action labels (moving, turning).
The two-stage approach—combining semi-supervised learning for partially labeled data with constrained loss for requirement satisfaction—demonstrates significant improvements in precision, recall, and F1-score compared to baseline models~\cite{moriyama2024mod}. Such integrated approaches bridge the gap between neural perception and symbolic reasoning, ensuring that perception outputs remain consistent with logical constraints.

\subsection{Uncertainty Quantification and Semantic Consistency}

Safe automated driving requires not only accurate perception but also reliable uncertainty estimation, especially in safety-critical scenarios. Recent methodologies have focused on incorporating domain knowledge into uncertainty quantification frameworks to enhance both statistical reliability and semantic consistency.

\subsubsection{Knowledge-Refined Prediction Sets}
Conformal prediction methods have emerged as powerful tools for uncertainty quantification in perception systems. These methods provide statistical guarantees on prediction sets, ensuring coverage of the true label with user-defined confidence levels. However, standard conformal prediction does not address semantic consistency across multiple perception tasks.
Knowledge-Refined Prediction Sets (KRPS) extend conformal prediction by ensuring that prediction sets across related perception tasks—such as agent classification, location classification, and action classification—are both statistically valid and semantically consistent \cite{doulaconformal}. This approach uses a knowledge graph to model semantic relationships between different tasks and employs a sequential set construction procedure that refines prediction sets based on an initial task.
Experimental results on ROAD and Waymo/ROAD++ datasets~\cite{giunchiglia2023road} demonstrate that KRPS reduces uncertainty by up to 80\% (producing smaller prediction sets) while increasing semantic consistency by up to 30\%, all while maintaining the coverage guarantees required by conformal prediction theory. This approach represents a significant advancement in making autonomous perception systems both more reliable through uncertainty quantification and more contextually aware through semantic consistency.

\subsubsection{Knowledge-Based Entity Prediction}

Perception systems often face challenges in detecting entities that may be missed due to hardware limitations, occlusions, or poor visibility. Knowledge-based Entity Prediction (KEP) addresses this challenge by leveraging structured knowledge representation and graph-based reasoning to infer potentially unrecognized entities based on context and scene understanding~\cite{wickramarachchi2021knowledge}.
KEP transforms scene data from autonomous driving datasets into knowledge graphs conformant with a dataset-agnostic Driving Scene Ontology. Through path reification and knowledge graph embeddings, KEP can predict missing entities with high precision (0.87 Hits@1) on real driving datasets such as PandaSet and NuScenes.
This neuro-symbolic approach demonstrates how perception systems can extend beyond traditional object detection to predict entities that should be in a scene based on context and knowledge, addressing a critical gap in current perception systems.

\subsection{Hierarchical Scene Understanding}

Autonomous vehicles must process and interpret complex traffic scenes at multiple levels of abstraction, similar to human drivers. The Hierarchical Knowledge-guided Traffic Scene Graph Representation Learning Framework (HKTSG) addresses this challenge by integrating both commonsense (domain-general) and expert-level (domain-specific) knowledge for improved traffic scene understanding~\cite{zhou2024hktsg, dickens2024mathematical}.

HKTSG constructs three types of scene graphs:
\begin{itemize}
    \item Global-Aware Graphs representing spatial and directional relationships among all traffic instances
    \item Pedestrian-Specific Graphs capturing motion patterns and behaviors of pedestrians
    \item Vehicle-Specific Graphs modeling interaction patterns between vehicles
\end{itemize}

This hierarchical approach, combined with relational graph attention and temporal transformers, enables the model to learn both local (intra-class) and global (inter-class) interactions, as well as spatio-temporal dependencies for dynamic traffic scenes. The integration of scene graph reasoning with global visual context improves both interpretability and completeness of perception.
Empirical results demonstrate significant performance gains, with HKTSG improving accuracy over CNN-LSTM~\cite{yurtsever2019risky} by 14.7\% on the IESG dataset and outperforming PCRA~\cite{liu2023learning} by 2-3\% on both IESG and Non-IESG datasets~\cite{liu2023learning}. These results highlight the benefits of structured, interpretable, hybrid models that blend symbolic scene understanding with deep learning.
\subsection{Verification and Validation of Perception Systems}
As perception systems incorporate more complex reasoning capabilities, ensuring their reliability becomes increasingly challenging. Recent advances in verification methodologies address this challenge by extending formal verification techniques to neural-symbolic systems.

\subsubsection{Scalable Neuro-Symbolic Verification}

Conventional verification methods struggle with the scalability of neuro-symbolic systems. To address this, Manginas et al. \cite{manginas2025scalable} propose a framework that compiles symbolic reasoning into arithmetic circuits and represents the full neuro-symbolic system as an end-to-end computational graph. Applying Interval Bound Propagation (IBP) to this graph enables approximate verification of whether small input perturbations affect the reasoning output. As exact verification is NP-hard~\cite{manginas2025scalable}, this relaxation-based approach prioritizes scalability over completeness. 

\subsubsection{Formal Methods for System Verification}

Formal methods, such as model checking and theorem proving, are widely used to ensure that autonomous vehicle decision models satisfy safety and legal requirements. Garcia et al.~\cite{garcia2021} provide a comprehensive overview of model checking techniques tailored for AV safety verification. For instance, König et al.~\cite{konig2024towards} apply temporal logic model checking to systematically verify that autonomous decisions adhere to formalized traffic rules. Beyond formal verification, high-fidelity simulation and real-world testing remain essential. Regulatory agencies such as the  Federal Highway Administration (FHWA) advocate for standardized testing frameworks in which legal compliance is a key performance metric. These integrated verification and validation pipelines combine sensor-level uncertainties with high-level decision checks to ensure that the entire perception-to-action stack aligns with legal and ethical standards.

\section{Formalizing Traffic Laws for Machine Reasoning}
\label{sec:formalization}
Representing traffic laws in a machine-interpretable form is a foundational step toward lawful AV behavior. In this section, we distinguish between two major lines of work: (A) logical formalization of driving rules, and (B) extracting formal specifications directly from natural language legal texts. This taxonomy reflects both theoretical encoding methods and practical extraction pipelines.
A critical first step towards legally aware autonomous driving is to represent traffic laws in a formal, machine-readable way. Traffic laws are written in natural language, making them difficult to implement directly in AV~\cite{Maierhofer2021Interstate}. Researchers have explored various methods to formalize these laws:

\subsection{Logical Specifications of Road Rules}
Temporal logics provide a mathematical framework for encoding time-dependent driving rules. Linear Temporal Logic (LTL)~\cite{rozier2011linear} and its variants can capture rules that involve temporal sequences (e.g., ``if the light is red, the car must eventually stop before the intersection”)~\cite{Gressenbuch2021Predictive}. For example, Esterle \emph{et al.} formalized a set of highway driving rules (e.g., keep a safe distance, pass on the left only) using LTL/MTL (Metric Temporal Logic)~\cite{Esterle2020TrafficRules}, providing precise mathematical interpretations of rules from the German traffic code. Similarly, Maierhofer \emph{et al.} formalized interstate driving rules based on German law and the Vienna Convention, translating provisions like safe following distance and overtaking restrictions into temporal logic formulas~\cite{Maierhofer2021Interstate}. This enables an automated check of whether a given trajectory or behavior sequence satisfies all the encoded rules. They later extended this approach to urban intersections, encoding complex right-of-way rules and traffic light regulations in MTL~\cite{Maierhofer2022Intersection}. Similarly,~\cite{alves2022towards} proposes a framework for creating a machine-readable Digital Highway Code using formal modeling and verification of timed automata to ensure autonomous vehicles adhere to the traffic rules.

One advantage of temporal logic (LTL, MTL, Signal Temporal Logic (STL)~\cite{donze2013signal}, etc.) is the ability to not only flag violations but also quantify \emph{robustness} of compliance. STL can assign a real-valued degree to which a rule is satisfied or violated~\cite{Gressenbuch2021Predictive}. This is useful for nuanced laws (like keeping a “sufficient” distance) by providing a margin. For instance, Arechiga used STL to formalize safety rules (like collision avoidance distances) and discussed using such specifications as contracts for evaluating driving software~\cite{Arechiga2019STL}.

Beyond temporal logics, other logic formalisms have been applied. Early works used deontic logic (the logic of obligations and permissions) to encode rules and exceptions—viewing laws as obligations that could sometimes be violated with justification~\cite{Costescu2017Accountable}. Higher-order logics and theorem proving have been employed to ensure rigor: Rizaldi \emph{et al.} formalized several traffic rules in the Isabelle/HOL proof assistant and even mechanically verified specific rules (like safe-distance keeping)~\cite{Rizaldi2017Isabelle}. Such proof-based approaches can guarantee that if the AV's model follows the formalized rules, specific undesirable outcomes will not occur under given assumptions. However, real-world driving often involves conflicting rules that cannot all be satisfied simultaneously. 

A notable concept addressing this challenge is the idea of \textbf{``rulebooks''} ~\cite{censi2019liability}. The rulebook is a hierarchy or priority ordering of rules. Since not all traffic rules are absolute, real driving may force choosing the “lesser evil” (e.g., crossing a line to avoid an obstacle)—rulebooks allow ranking rules by importance. For example, a rulebook might specify that avoiding collisions has higher priority than staying perfectly centered in the lane. The rulebook framework provides a structured way to resolve conflicts. If a low-priority rule must be broken to satisfy a higher-priority rule, the system can do so in a principled manner. Xiao \emph{et al.} formalized this via a Total Order over Equivalence Classes (TORQ) of rules, ensuring every rule is comparable by priority~\cite{Xiao2021Optimal}. This formalization extends earlier ideas of “least violating” control synthesis~\cite{Tumova2013LeastViolating}  by systematically relaxing lower-priority rules when necessary.

\subsection{From Legal Text to Machine-Readable Rules}
Bridging the gap from human legal text to formal specs is a research challenge. McLachlan \emph{et al.} proposed a pipeline to \textit{deconstruct} traffic law into formal representations~\cite{McLachlan2022Deconstructing}. They translate legal text into a structured English logic, then into Boolean logic, creating what they call “Lawmaps” to visualize the flow of legal requirements. Using this, they demonstrated a Bayesian network that captures probabilistic relationships of rule compliance and exceptions, intended to be understandable by both engineers and legal experts. Such interdisciplinary efforts help ensure that the formal rules correctly capture the intent of the law (with validation by legal experts) and that any assumptions or thresholds introduced are legally defensible.

More recently, large-scale efforts have begun to \textit{digitize traffic regulations} into databases. For example, the US Federal Highway Administration in 2021 started developing a prototype regulation database for AVs~\cite{Han2025RegulationAware}. Shi and Wang built a modular framework that takes digitized traffic rules and uses them to guide trajectory planning~\cite{Shi2025Digitizing}  (we will discuss the planning aspect in Section \ref{sec:planning}). The digitization involves encoding rules from statutes and driver manuals into machine-interpretable formats (logic formulas or structured data entries). This ensures comprehensive coverage of rules and eases updating as laws change. However, encoding an entire traffic code is arduous—researchers often start with a subset (e.g., rules for highway driving or basic intersection protocols) as seen in many works.

\section{Legal Compliance in Motion Planning}
\label{sec:planning}

Motion planning for autonomous vehicles (AVs) typically seeks to generate safe, efficient, and comfortable trajectories. Ensuring legal compliance adds additional complexity, as planners must now incorporate traffic rules as formal constraints or allow principled trade-offs when strict adherence conflicts with safety. In this section, we group the existing strategies into rule-constrained trajectory optimization and runtime verification of generated plans.

A direct approach to legal compliance is to encode traffic rules as constraints within a trajectory optimization problem. For example, planned speeds must not exceed legal limits, and trajectories should avoid crossing solid lane markings. STL-based specifications~\cite{Arechiga2019STL} serve this purpose by expressing rules as temporal logic constraints or cost functions in optimal control solvers. If a rule can be written as a mathematical inequality—such as maintaining a safe following distance $d \geq d_{\text{min}}$—it can be directly enforced. When rules are soft (i.e., violable under certain conditions), penalties can be incorporated into the cost function to prioritize among competing objectives~\cite{Tumova2013LeastViolating}.

\subsection{Rule-Constrained Trajectory Optimization}

Autonomous vehicle planning stacks are often hierarchical, where a high-level behavior planner selects maneuvers (e.g., “change lane to the left”) and a low-level motion planner computes the precise trajectory. Legal compliance can be integrated at the decision-making level by restricting available maneuvers. For example, finite-state machines (FSMs) can model driving states such as lane-keeping, turning, or yielding. Han~\emph{et al.}~\cite{Han2025RegulationAware} extend an FSM-based planner with a regulation database. Their system automatically retrieves relevant traffic rules based on scene understanding—using vision and language processing—and removes maneuvers that violate those rules. For instance, if a “no passing” zone is detected, the FSM disables lane-change maneuvers for overtaking. This ensures legal compliance by filtering behavior options through rule-based constraints.

Recent work has explored using large language models (LLMs) to support legal reasoning in planning. Cai~\emph{et al.}~\cite{Cai2024DrivingRegulation} proposed a system where a regulation retrieval module identifies applicable traffic laws, and an LLM interprets them to advise the planner. For example, when approaching a school bus with flashing lights, the system fetches the relevant law and the LLM reasons that the AV should stop. This method handles regional rule variability and distinguishes between hard laws and advisory guidelines, outputting a natural language explanation alongside the decision. Although experimental, it suggests that LLMs can serve as interpretable legal advisors, similar to how human drivers consult manuals.

\textbf{Rule-based planners} also remain relevant due to their interpretability. These systems follow explicit, human-readable rules—for example, ``if the light is red, decelerate to a stop.” While such systems once fell out of favor due to rigidity, hybrid approaches are reviving them with formal underpinnings. Thornton~\emph{et al.}~\cite{Thornton2017EthicalControl} developed an “ethical controller” that encodes legal principles (e.g., yielding) into the control law. The controller overrides nominal plans when it detects a rule or ethical violation, acting as a built-in legal safeguard.

\subsection{Verification and Runtime Monitoring in Planning}
Beyond generating a plan, an AV system should verify that the planned trajectory complies with applicable legal rules before execution. Formal verification techniques are increasingly used to check whether planned actions satisfy such specifications. Runtime verification monitors, for example, can observe the executed trajectory in real-time and flag violations as they occur or are imminent~\cite{Gressenbuch2021Predictive}. Many planning systems now include a monitoring subsystem that evaluates formal rule formulas (e.g., ``always stay within speed limit”) against the planned trajectory. If a violation is detected, the plan is rejected, or a safety supervisor intervenes.

Rizaldi~\emph{et al.}~\cite{Rizaldi2017Isabelle} developed formally verified rule checkers where even the monitoring logic is proven correct, ensuring no violation is missed due to implementation bugs. In practice, more lightweight implementations are standard. For instance, Yu~\emph{et al.}~\cite{Yu2024LegalMonitor} present a monitor that continuously checks inter-vehicle distances and speed limits and can trigger emergency braking or trajectory overrides when thresholds are crossed.

In summary, legal compliance in planning can be supported through formal verification and runtime monitoring. These methods provide a reactive safety layer that complements proactive rule filtering. A recurring theme is that if a rule must be violated due to external circumstances, the system should do so minimally, justifiably, and ideally with an auditable explanation. We now turn to the prediction module, which introduces a different but related set of challenges for integrating legal norms.

\section{Handling Ambiguities and Exceptions in Laws}
\label{sec:ambiguity}
Traffic laws are written for human interpretation and often contain ambiguity, vagueness, and exceptions resolved through social norms or judicial discretion. Autonomous vehicles must therefore do more than follow formalized rules; they must understand context, balance competing priorities, and justify necessary deviations. In this section, we explore how ambiguity, exceptions, and jurisdictional variability are addressed in current AV systems and how interpretability mechanisms ensure these decisions remain transparent and accountable.

\subsection{Norm-Aware Behavior Prediction}
\label{sec:prediction}
Predicting the behavior of other traffic participants (vehicles, pedestrians, cyclists) is a key function of an AV’s decision-making system. Legal reasoning can inform prediction in two main ways: by assuming that others will follow the rules (to reduce uncertainty and narrow the prediction space), and by detecting or forecasting when others might \emph{not} follow the rules (to anticipate dangerous situations).

Many traditional prediction models implicitly assume normal driving behavior, which includes general compliance with traffic laws. For example, a physics-based prediction might assume a car will decelerate when approaching a red light or stop line, because that’s what lawful drivers do. Data-driven prediction models, trained on human driving data, inherently reflect the frequency of rule obedience vs. violations in that data (most drivers stop at red lights nearly 100\% of the time, so a learned model will predict stopping). However, to robustly handle edge cases, AV practitioners have to encode traffic rules explicitly into prediction algorithms.

Gressenbuch and Althoff tackled the flipside: how to predict \emph{rule violations} by others~\cite{Gressenbuch2021Predictive}. They observed that AV planners rely on predictions that usually assume compliance, so if another driver is going to do something illegal (say, run a stop sign), a naive AV might not be prepared. They formalized traffic rules in temporal logic (as described earlier). Then, they trained a neural network to classify if an observed agent was likely to violate a rule shortly. By extracting features from the quantitative satisfaction values of the rule formulas, their system could learn telltale signs (e.g., a car approaching an intersection at high speed might violate the stop requirement). Experiments on highway data showed improved prediction of events like illegal lane changes. This kind of \textbf{predictive monitoring} combines rule-based logic with learning to handle the inherently low-probability but high-risk event of another agent breaking the law.

The environmental context often indicates which rules apply, and prediction models can be conditioned on that context. For example, if there is a “no U-turn” sign, a well-informed predictor would assign essentially zero probability to another car making a U-turn there (assuming drivers follow the posted rule). Incorporating such context requires the AV to perceive traffic signs/markings and feed that into the prediction system. Han \emph{et al.}’s vision-language approach~\cite{Han2025RegulationAware}, while primarily used for planning, could likewise feed a prediction model: the recognized regulation (e.g., “yield to pedestrians”) could inform that the oncoming vehicle is likely to slow down (since it must yield if pedestrians are present). 

There is also a safety angle: by predicting possible violations by others, the AV can plan more defensively. If our prediction system flags that the car behind us is approaching fast and might not obey the speed limit or might ignore that we are stopping, the AV could choose to change lanes or get out of the way preemptively. This is essentially \textbf{prediction as a defensive driving tool}. Huang \emph{et al.}’s reinforcement learning planner with a predictive critic~\cite{Huang2025PredictiveRL} embodies this, where the planner adjusts now to avoid the situation of someone else’s future violation.

In summary, integrating legal norms into prediction is about encoding the assumption of normal, law-abiding behavior (to simplify predictions and make them realistic), and simultaneously being alert to anomalies that indicate rule-breaking. Formal rule models and learned classifiers for violations are complementary: the former defines what counts as a violation, and the latter deals with the uncertainty of if/when a human will commit one. By combining these, AVs can better anticipate the actions of others in traffic.

\subsection{Ambiguity in Natural Language Laws}

A rule like ``maintain a safe distance” raises the practical question: how safe is safe? Laws may provide approximate thresholds (e.g., ``at least 2 seconds”), but these are interpreted flexibly in practice. Yu \emph{et al.}~\cite{Yu2024LegalMonitor} propose calibrating such thresholds using statistical norms extracted from large-scale human driving data. For example, by analyzing hundreds of lane-change cases, they determine the lateral deviation that typically constitutes interference, making the rule operational for monitoring compliance. This approach mirrors how human drivers—and courts—evaluate reasonableness: not from absolute thresholds, but from socially normative behavior.

Speed limits offer another gray area. While humans often exceed the limit by small margins with no legal consequences, AVs might be programmed to obey them strictly. McLachlan \emph{et al.}~\cite{McLachlan2022Deconstructing} incorporate a 5\% tolerance to match real-world enforcement. These tuning decisions reflect a key dilemma: strict rule-following may harm flow or be perceived as overly cautious, while deviation introduces legal risk.

Formal methods papers often mention assumptions or interpretations made. For instance, Maierhofer \emph{et al.}~\cite{Maierhofer2021Interstate} concretized the German law that requires a “sufficient distance” by incorporating court interpretations and standard formulas (like the “half the speedometer reading in meters” rule-of-thumb in Germany). They worked with legal experts so that their logical rules align with judicial precedent—an essential step because the letter of the law might be vague, but case law provides concrete guidance.

\subsection{Rule Exceptions and Context}
When can an AV break traffic rules? This question is central to developing legally compliant autonomous systems that can handle emergencies. Researchers are actively studying how to encode exceptions and higher-level principles that permit violations in extreme cases. Manas and Paschke developed a framework that explicitly models \textbf{rule exceptions}~\cite{Manas2023FormalizedExceptions}. They formalize traffic rules in a hierarchical structure with base rules and layers of exceptions. Consider a concrete example: the base rule states ``do not cross a solid line," but an exception layer permits ``crossing a solid line to bypass a stalled vehicle if done with caution." Using temporal logic with additional predicates, their system represents the rule and the conditions under which it can be overridden. This approach enables monitoring beyond simple violations—the system can determine whether a violation is \emph{justified} due to an exception condition. For accountability purposes, if an AV performs an illegal maneuver to avoid an accident, the system logs that the maneuver, while breaking a rule, was legally permissible under an emergency exception. This nuance moves autonomous systems closer to legal reasoning rather than mere rule-following.

The concept of \textbf{necessary violation} appears in the literature on ethical decision-making for AVs~\cite{Thornton2017EthicalControl}. In unavoidable crash scenarios, an AV might need to violate a traffic law (such as swerving into the wrong lane) to minimize harm. From a legal perspective, human drivers might invoke the doctrine of necessity—a legal defense that justifies rule violations to prevent greater harm. Some research has attempted to encode such principles so that AVs can choose a lesser violation when no entirely lawful action exists.

Censi et al. introduced a hierarchy extending beyond laws to include broader categories: liability, ethics, and cultural norms~\cite{censi2019liability}. In their rulebook concept, traffic laws have high priority (since violating them carries legal consequences). Still, ethical considerations (such as not harming pedestrians) rank even higher, while "culture-aware" conventions (like zipper merging etiquette) rank lower. Such structured rule hierarchies inherently handle exceptions: when an ethical rule conflicts with a legal rule, the ethical rule takes precedence, effectively creating an exception to the legal rule. This provides a rational framework for rule violations: the system can later explain it as obeying a higher imperative.

\subsection{Multi-Jurisdiction and Adaptability}
Ambiguity also arises when considering different jurisdictions, where the same legal concept may be defined or enforced differently, introducing uncertainty and variability in interpretation across regions. A major practical challenge is making the AV’s legal reasoning \emph{adaptable}. Frameworks that rely on centralized rule databases or retrieval (like Han \emph{et al.}~\cite{Han2025RegulationAware} or Cai \emph{et al.}~\cite{Cai2024DrivingRegulation}) aim to address this by swapping out the source of rules while keeping the decision logic unchanged.

Using an LLM to interpret rules on the fly is promising in this regard, as it could allow one core model to drive in many countries by feeding it local codes and signage. However, verifying that the interpretation is correct is especially challenging. LLMs may hallucinate rules, misinterpret exceptions, or yield inconsistent conclusions—unacceptable issues in a safety-critical, legally bound system.

As a safer alternative, researchers have proposed structured-English-to-MTL pipelines~\cite{manastr2mtl}, where legal experts author rules in a constrained syntax that can be deterministically compiled to formal logic, balancing adaptability with verifiability.    

In conclusion, handling ambiguities and exceptions requires blending legal domain knowledge with technical solutions. Data-driven tuning of parameters, multi-tier logic representations, and hierarchy of rules help an AV not just follow laws, but follow them in a way that aligns with human interpretation and common sense. It’s not enough to be binary lawful/unlawful; the AV must handle the gray areas similarly to a reasonable human or court. In the next section, we look at how the decisions made under these frameworks can be explained and audited.

\subsection{Legal Interpretability and Explainability}
\label{sec:explain}
When an autonomous vehicle is involved in an incident or traffic stop, explaining its actions in legal terms becomes crucial for liability determination and public trust~\cite{Cai2024DrivingRegulation}. This requirement has driven research toward transparent and interpretable AV decision-making systems that can justify their actions concerning traffic laws and regulations.

Several approaches discussed earlier inherently provide \textbf{explanations} for legal compliance decisions. Cai \emph{et al.}'s LLM-based reasoning module generates natural language explanations linking specific rules to chosen actions~\cite{Cai2024DrivingRegulation}. For instance, their system outputs: ``It is illegal to turn right on red at this intersection because a sign prohibits it, so the vehicle waits." Such explanations can be communicated to passengers or authorities to build trust and demonstrate compliance. Rule-based and logic-based systems naturally support explainability because their decision processes center on human-understandable legal concepts.

Formal verification approaches contribute to interpretability through \textbf{proofs and counterexamples}. When a planning algorithm cannot satisfy all applicable rules simultaneously, it can identify which rules conflict in the current scenario, explaining why a violation became necessary. This capability supports the development of \emph{legal safety cases} for AVs—structured arguments demonstrating that the vehicle violates laws only under justifiable circumstances. Each deviation from nominal behavior can be documented with hierarchical justifications (e.g., ``violated rule A to satisfy higher-priority rule B").

Visual representations also enhance legal interpretability. McLachlan \emph{et al.}'s ``Lawmaps" diagram the logical structure of rules and their exceptions, making decision flows comprehensible to both engineers and legal experts~\cite{McLachlan2022Deconstructing}. Such documentation can support certification processes and facilitate incident analysis by clearly showing how the AV's legal reasoning policies operate.

\textbf{Post-incident accountability} represents a critical interpretability requirement. Following collisions, investigators must determine whether the AV operated legally and understand any violations. Yu \emph{et al.}'s online compliance monitoring system continuously evaluates AV behavior against traffic laws, creating detailed records for incident reconstruction~\cite{Yu2024LegalMonitor}. Their work emphasizes that independent monitoring provides ``substantial evidence for the traceability of traffic incidents." For example, their monitor might log: ``vehicle exceeded lane boundary by 0.2m to avoid obstacle—within acceptable emergency deviation." Such records can demonstrate in court that the AV made minimal, legally justified deviations to comply with higher-priority safety requirements.

Beyond post-incident analysis, interpretability extends to real-time communication with other road users. When an AV must perform unexpected but legally justified maneuvers, explicit signaling becomes important since humans may not anticipate autonomous system behavior. For instance, an AV might use external displays to communicate "Yielding to emergency vehicle" when making an otherwise illegal lane change. While human-vehicle communication falls outside our surveyed literature, it represents a logical extension of transparent legal decision-making in autonomous systems.

\section{Discussion and Open Challenges}
\label{sec:discussion}
Embedding legal specifications into AV prediction and planning has seen substantial progress, but many challenges remain:

\subsubsection{Completeness and Scalability} Most research to date tackles a subset of traffic rules (highway driving, basic traffic signals, etc.). Encoding the \emph{full} traffic code of even one jurisdiction, and maintaining it as laws update, is daunting. Approaches like LLM-based retrieval~\cite{Cai2024DrivingRegulation} offer scalability by handling raw text, but ensuring 100\% correct interpretation is unresolved. Tools for semi-automatically translating entire legal texts into formal rules (with a human in the loop) will be needed to scale up.

\subsubsection{Validation Across Scenarios} A rule-based system might work well in the scenarios it was designed for, but strange combinations of events could expose gaps or conflicts. How do we guarantee an AV will handle a rare situation where multiple obscure rules interact? Formal methods offer some guarantees (e.g., model checking all combinations within some bounds), but state-space explosion is an issue. Scenario generation tools, like falsification testing in control~\cite{Mehdipour2023FormalMethodsSurvey}, can help find where the legal logic might break down.

\subsubsection{Learning and Adaptation}
Driving norms vary, and overly strict AVs risk disrupting traffic or being exploited by aggressive drivers. Yet, adapting to human behavior can lead to non-compliance. Striking a balance is challenging. Some suggest AVs adopt "socially acceptable" behaviors, like mild speeding or rolling stops, but legal obligations remain. This tension between lawful and natural driving is unresolved, and future policies may grant AVs limited flexibility. For now, stakeholders have to navigate this delicate trade-off—an open challenge in both research and regulation.

\subsubsection{Cross-Jurisdiction Operations} If an AV drives from one country or state to another, it must seamlessly switch its rule set. Misinterpretation could be dangerous (e.g., right turn on red is legal in most US states but illegal in some countries). Standardizing machine-readable traffic rules (similar to how digital maps are standardized) could greatly aid this. The aforementioned FHWA regulation data framework is a start~\cite{Han2025RegulationAware}. International efforts might be needed to define core rule ontologies that each locale can extend.

\subsubsection{Human Factors and Communication} As AVs that strictly follow rules mix with human drivers who sometimes don’t, how should the AV anticipate and react? We covered the prediction of violations, but what about influencing others? Suppose an AV never nudges forward at a four-way stop. In that case, it might sit forever because humans expect some assertiveness (technically, creeping into an intersection is sometimes a violation if done too early). Should the AV break a minor rule to signal intent? These borderline behaviors tie into game theory and may require new traffic laws to accommodate AV behavior.

\subsubsection{Regulatory Acceptance and Legal Liability} 
Ultimately, a court or regulator will judge an AV’s actions. The industry might push for a “reasonable AV standard” in law that acknowledges some flexibility. In the interim, proving that an AV’s planning module verifiably complies with all laws (except emergencies) could reduce liability. Methods like formal verification and certified monitoring~\cite{Rizaldi2017Isabelle}  might become part of the safety case for an AV. An open challenge is how to certify an LLM-based reasoning system – these are not yet amenable to formal verification, so one might need to constrain or heavily test them.

\vspace{-2pt}
\section{Conclusion}

This survey has detailed the burgeoning field of integrating legal specifications into AV decision-making. From formal logics that capture the letter of traffic laws, to hierarchical rulebooks that encode the spirit of the law with priorities, and from rule-constrained trajectory planners to violation-aware predictors, the research community is laying the groundwork for legally-savvy AVs. We have seen that while hard-coding traffic rules using formal logic enables rigorous compliance, it is often insufficient by itself to handle real-world ambiguities, exceptions, and distributional shifts. Hybrid approaches that combine formal methods with learning-based techniques are increasingly necessary.

Notably, making an AV law-compliant is not just a technical goal, but a sociotechnical one: it involves translating human norms into the AV stack and ensuring the outcomes are acceptable to the public and legal authorities. As AVs advance to widespread deployment, we expect increased collaboration between roboticists, legal scholars, and regulators to refine these approaches. Areas like explainable AI and formal verification are key to demonstrating legally reasonable AV behavior.

\vspace{-3pt}
\section*{Acknowledgment}
\vspace{-6pt}
The research is funded by the German Federal Ministry for Economic Affairs and Energy within the project ''NXT GEN AI METHODS – Generative Methoden für Perzeption, Prädiktion und Planung''.

{\scriptsize
\bibliographystyle{IEEEtran}
\bibliography{main}
}

\end{document}